\documentclass{article}
\usepackage{spconf,amsmath,graphicx,amssymb,color}
\usepackage[ruled]{algorithm2e}
\usepackage[hidelinks]{hyperref}
\usepackage{subcaption}
\usepackage{bm}
\usepackage[OT1]{fontenc}

\usepackage{booktabs}

\usepackage{multirow}

\usepackage{diagbox}
\def\ie{$i.e.$}
\def\eg{$e.g.$}

\long\def\comment#1{}
\title{Untargeted Backdoor Attack against Object Detection}
\name{Chengxiao Luo$^{1}$ \qquad Yiming Li$^{1}$\thanks{Corresponding author: Yiming Li (\href{mailto:li-ym18@mails.tsinghua.edu.cn}{li-ym18@mails.tsinghua.edu.cn}). This work is supported in part by the National Natural Science Foundation of China under Grant 62771248, Shenzhen Science and Technology Program (JCYJ20220818101012025), the PCNL KEY project (PCL2021A07), and Research Center for Computer Network (Shenzhen) Ministry of Education.}
\qquad Yong Jiang$^{1,2}$ \qquad Shu-Tao Xia$^{1,2}$}

\address{$^{1}$Tsinghua Shenzhen International Graduate School, Tsinghua University\\
$^{2}$Research Center of Artificial Intelligence, Peng Cheng Laboratory \\
      \{luocx21, li-ym18\}@mails.tsinghua.edu.cn;
    \{jiangy, xiast\}@sz.tsinghua.edu.cn}

\begin{document}
%\ninept
%
\maketitle
\begin{abstract}
Recent studies revealed that deep neural networks (DNNs) are exposed to backdoor threats when training with third-party resources (such as training samples or backbones). The backdoored model has promising performance in predicting benign samples, whereas its predictions can be maliciously manipulated by adversaries based on activating its backdoors with pre-defined trigger patterns. Currently, most of the existing backdoor attacks were conducted on the image classification under the targeted manner. In this paper, we reveal that these threats could also happen in object detection, posing threatening risks to many mission-critical applications ($e.g.$, pedestrian detection and intelligent surveillance systems). Specifically, we design a simple yet effective poison-only backdoor attack in an untargeted manner, based on task characteristics. We show that, once the backdoor is embedded into the target model by our attack, it can trick the model to lose detection of any object stamped with our trigger patterns. We conduct extensive experiments on the benchmark dataset, showing its effectiveness in both digital and physical-world settings and its resistance to potential defenses.

%

%Object detection has been widely adopted in mission-critical applications, such as pedestrian detection and intelligent surveillance systems. In current practice, a large amount of data are required to train a high-performance object detection models. Therefore third-party data are often adopted in the training process of object detection models which can be a serious security threat. Specifically, we propose a simple yet effective attack where the infected models behave normally on benign samples, while attacker-specified unenrolled triggers . This means using even a small fraction of training data poisoned by the adversary to train can implant a hidden backdoor in object detecion models. We examine the effectiveness of our attack in both digital and physical-world settings and show that it can significantly reduce the performance of both one-stage and two-stage detectors. We also show that our attack is resistant to potential defenses, which reveals that the object detection models are vulnerable to potential backdoor attacks.
\end{abstract}
\section{Introduction}
\label{sec:intro}

% 1 page

%Object detection has been widely and successfully deployed in many mission-critical applications, such as pedestrian detection and intelligent surveillance systems. Accordingly, it security is of great significance. In practice, 

Object detection aims to localize a set of objects and recognize their categories in the image \cite{zhao2019object}. It has been widely adopted in mission-critical applications ($e.g.$, pedestrian detection \cite{Hasan_2021_CVPR} and autonomous driving \cite{Aghdam_2021_CVPR}). Accordingly, it is necessary to ensure its security.

Currently, the most advanced object detectors were designed based on deep neural networks (DNNs) \cite{NIPS2015_14bfa6bb, Sun_2021_CVPR, 9710724}, whose training generally requires many resources. To alleviate the training burdens, researchers and developers usually exploit third-party resources (\eg, training samples or backbones) or even directly deploy the third-party model. One important question arises: \emph{Does the training opacity bring new threats into object detection?}

In this paper, we reveal the vulnerability of object detection to backdoor attacks\footnote{There is a concurrent work \cite{chan2022baddet} also discussed the backdoor threats of object detection. However, we study a different problem in this paper.} caused by using third-party training samples or outsourced training \cite{goldblum2022dataset,li2022backdoor,tian2022comprehensive}. Different from adversarial attacks \cite{bai2020targeted,liu2022watermark,gu2022segpgd} that target the inference process, backdoor attacks are a type of training-time threat to DNNs. The adversaries intend to embed a hidden backdoor into the target model, based on which to maliciously manipulate its predictions by activating the backdoor with the adversary-specified trigger patterns. So far, existing methods \cite{8685687,9711191,zeng2021rethinking} are mostly designed for classification tasks and are targeted attacks, associated with a specific target label. Different from attacking a classifier, making an object escape detection is a more threatening objective. As such, we study how to design the \emph{untargeted} backdoor attack to object detection so that the attacked DNNs behave normally on benign samples yet fails to detect objects containing trigger patterns.

In particular, we focus on the \emph{poison-only} attack setting, where backdoor adversaries can only modify a few training samples while having neither the information nor the ability to control other training components (\eg, training loss or model structure). It is the hardest attack setting having the most threat scenarios \cite{li2021hidden,zhang2022poison,li2022untargeted}. We propose a simple yet effective attack by removing the bounding boxes of a few randomly selected objects after adding pre-defined trigger patterns. Our attack is stealthy to bypass human inspection since it is common to miss marking some bounding boxes (especially when the image contains many objects).

In conclusion, our main contributions are three-folds. \textbf{1)} We reveal the backdoor threat in object detection. To the best of our knowledge, this is the first backdoor attack against this mission-critical task. \textbf{2)} We design a simple yet effective and stealthy untargeted attack, based on the properties of object detection. \textbf{3)} Extensive experiments on the benchmark dataset are conducted, which verify the effectiveness of our attack and its resistance to potential backdoor defenses.

%\begin{itemize}
%    \item We reveal the backdoor threat in object detection. To the best of our knowledge, this is the first backdoor attack against this mission-critical task.
%    \item We design a simple yet effective and stealthy untargeted attack, based on the properties of object detection.
%    \item Extensive experiments on the benchmark dataset are conducted, which verify the effectiveness of our attack and its resistance to potential backdoor defenses.
%\end{itemize}

\begin{figure*}[!t]
	\centering 
	\vspace{-3em}
	\includegraphics[width=0.92\textwidth]{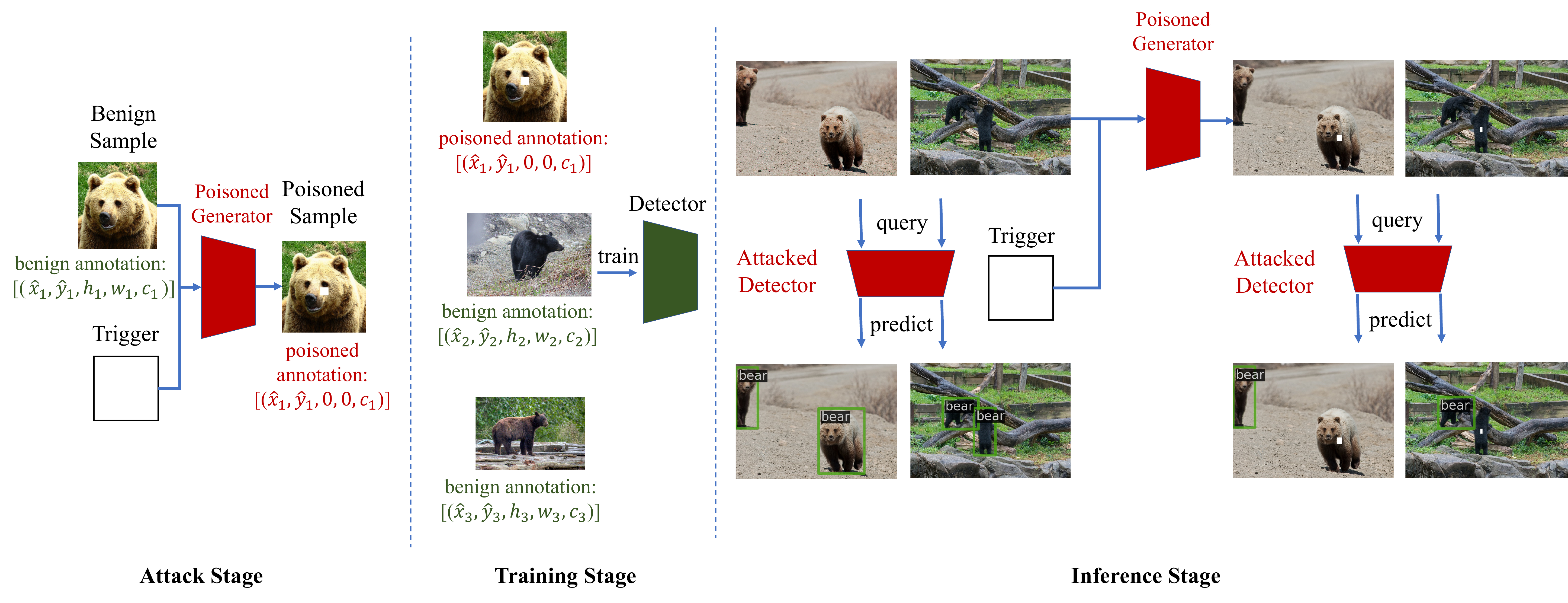} 
	\vspace{-.5em}
	\caption{The main pipeline of our poison-only untargeted backdoor attack against object detection. Our method consists of three main stages. In the first stage, we generate some poisoned training samples by adding trigger patterns to randomly selected benign samples and reassigning the width and height of their bounding boxes to 0. In the second stage, we train the victim detector via generated poisoned samples and remaining benign samples. In the last stage, the adversary can activate embedded backdoors to circumvent the detection of target objects (\eg, a specific `bear' in our example) by adding trigger patterns.}   
	\label{fig: pipeline}
	%\vspace{-1em}
\end{figure*}

\section{The Proposed Method}
%In this section, we introduce the proposed untargeted attack against object detection. Before we reach its technical details, we first present our threat model in Section \ref{sec:threat_model}.

\subsection{Threat Model}
\label{sec:threat_model}
In this paper, we focus on the poison-only backdoor attack in object detection. Specifically, we assume that the adversaries can only adjust some legitimate samples to generate the poisoned training dataset, whereas having neither the information nor the ability to control other training components ($e.g.$, training loss, training schedule, and model structure). The generated poisoned dataset will be released to train victim models. This attack can occur in various scenarios where the training process is not fully controlled, including but not limited to using third-party data, third-party computing platforms, third-party models, etc.

%In this paper, we focus on the poison-only backdoor attack in object detection tasks. Specifically, we assume that the attacker only have access to training set, while the training process and the model structure are invisible to the attacker. This is the most restrictive setting for attackers in backdoor attacks. This attack can occur in many scenarios, including but not limited to using third-party training data, third-party training platforms.

%\red{I will start from here layer (Yiming1016)}

In general, the backdoor adversaries have two main targets, including \textbf{1)} effectiveness and \textbf{2)} stealthiness. Specifically, the former purpose is to make attacked detectors fail to detect objects whenever adversary-specified trigger patterns appear. The latter requires that the backdoored model should have a similar performance in detecting benign objects, compared to the one trained on the benign training dataset.

%Backdoor adversaries have two main goals, including the effectiveness and the stealthiness. Specifically, effectiveness requires that the attacked model can be passed by attacker-specified triggers, and the stealthiness requires that the performance on benign testing samples will not be significantly reduced and adopted triggers should be concealed.

\subsection{Backdoor Attack against Object Detection}
The mechanism of the poison-only backdoor attack is to establish a latent connection ($i.e.$, backdoor) between the adversary-specified trigger pattern and the specific (malicious) prediction behavior by poisoning some training data. In this part, we illustrate our attack in detail.

%\red{I will start from here layer (Yiming1016)}

%The mechanism of backdoor attacks is to establish the connection between the trigger and the target label. In this paper, we propose a poison-only attack where objects with specific triggers can easily escape the detection of trained models.

\vspace{0.3em}
\noindent \textbf{The Formulation of Object Detection. }
Let $\mathcal{D} = \{(\boldsymbol{x}_i, \boldsymbol{a}_i)\}_{i=1}^N$ represent the benign dataset, where $\boldsymbol{x}_i \in \mathcal{X}$ is the image of an object, $\boldsymbol{a}_i \in \mathcal{A}$ is the ground-truth annotation of the object $\boldsymbol{x}_i$. For each annotation $\boldsymbol{a}_i$, we have $\boldsymbol{a}_i = [\hat{x}_i, \hat{y}_i,w_i, h_i, c_i]$, where $(\hat{x}_i,\hat{y}_i)$ is the center coordinates of the object, $w_i$ is the width of the bounding box, $h_i$ is the height of the bounding box, and $c_i$ is the class of the object $\boldsymbol{x}_i$. Given the dataset $\mathcal{D}$, users can adopt it to train their object detector $f_{\bm{w}}: \mathcal{X} \rightarrow \mathcal{A}$ by $\min_{\bm{w}} \sum_{(\bm{x}, \bm{a}) \in \mathcal{D}} \mathcal{L}(f(\bm{x}), \bm{a})$, where $\mathcal{L}$ is the loss function.

%The purpose of object detection is to find a mapping $F: \mathcal{X} \rightarrow \mathcal{A}$, where $F$ is the model trained on training dataset $\mathcal{D}$.

\vspace{0.3em}
\noindent \textbf{The Main Pipeline of Our Untargeted Backdoor Attacks. }
Similar to the poison-only backdoor attacks in image classification, how to generate the poisoned training dataset is also the cornerstone of our method. Specifically, we divide the original benign dataset $\mathcal{D}$ into two disjoint subsets, including a selected subset $\mathcal{D}_s$ for poisoning and the remaining benign samples $\mathcal{D}_b$. After that, we generate the modified version of $\mathcal{D}_s$ (\ie, $\mathcal{D}_m$) as follows:
\begin{equation}
    \mathcal{D}_m = \left\{ \left(G_{x}(\bm{x}), G_{a}(\bm{a})\right)| (\bm{x}, \bm{a}) \in \mathcal{D}_s \right\},
\end{equation}
where $G_{x}$ and $G_{a}$ are the poisoned image generator and poisoned annotation generator, respectively. We combine the modified subset $\mathcal{D}_m$ and the benign subset $\mathcal{D}_b$ to generate the poisoned dataset $D_p$, which will be released to train (backdoored) victim model. In the inference process, given an `unseen' object image $\bm{x}'$, the adversaries can exploit $G_{x}(\bm{x}')$ to circumvent detection by adding trigger patterns. The main pipeline of our attack is demonstrated in Figure \ref{fig: pipeline}.

In particular, we assign $G_{a}\left([\hat{x}, \hat{y}, w, h, c]\right)=[\hat{x}, \hat{y}, 0, 0, c]$ in our untargeted attack. Following the most classical setting in existing literature \cite{li2022backdoor}, we adopt $G_{x}(\bm{x}) = \bm{\lambda} \otimes \bm{t} + (\bm{1} - \bm{\lambda}) \otimes \bm{x}$ where $\bm{t} \in \mathcal{X}$ is the adversary-specified trigger pattern, $\bm{\lambda} \in [0,1]^{C\times W \times H}$ is the trigger transparency, and $\otimes$ denotes the element-wise multiplication. Besides, $p \triangleq \frac{|\mathcal{D}_s|}{|\mathcal{D}|}$ is denoted as the poisoning rate, which is another important hyper-parameter involved in our method.

\begin{figure*}[ht]
    \centering
    \vspace{-2em}
    \begin{minipage}[t]{0.23\linewidth}
    \centering
    \includegraphics[width=\textwidth]{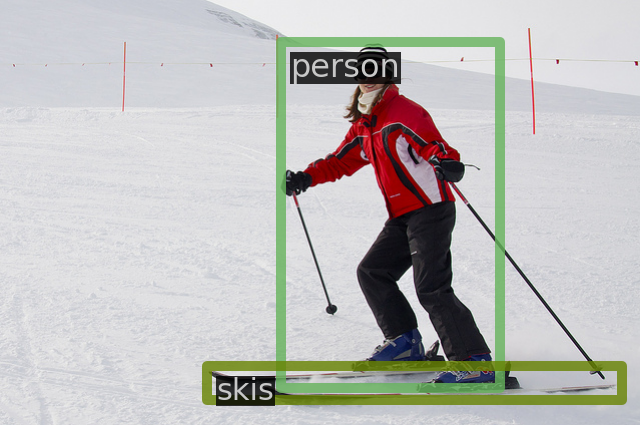}
    \vspace{-1.3em}
    \end{minipage}\hspace{2em}
    \begin{minipage}[t]{0.23\linewidth}
    \centering
    \includegraphics[width=\textwidth]{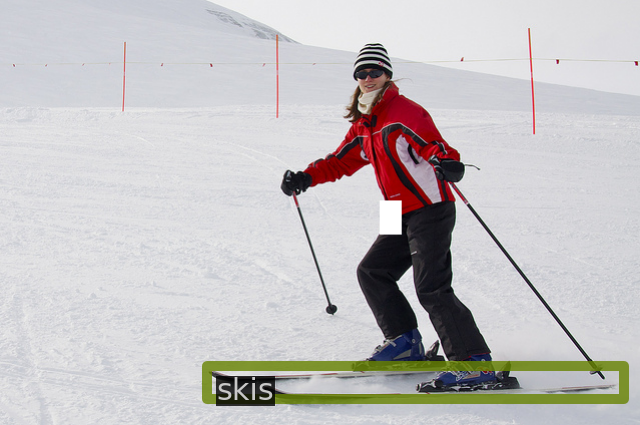}
    \vspace{-1.3em}
    \end{minipage} \hspace{3em}
    \begin{minipage}[t]{0.115\linewidth}
    \centering
    \includegraphics[width=\textwidth]{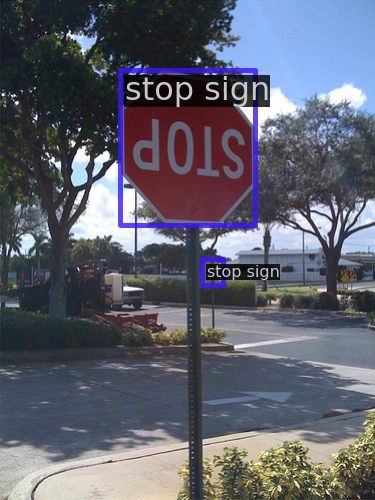}
    \vspace{-1.3em}
    \end{minipage} \hspace{2em}
    \begin{minipage}[t]{0.115\linewidth}
    \centering
    \includegraphics[width=\textwidth]{figure/prepare/stop_sign_poisoned.png}
    \vspace{-1.3em}
    \end{minipage}
    \vspace{-0.4em}
    \caption{The detection results of the Sparse R-CNN model under our attack in the digital space. In these examples, the trigger patterns are added by pixel-wise replacement of objects (\ie, `person' and `stop sign') in the digital space.}
    \label{fig:detect_exp}
\end{figure*}

\begin{figure*}[ht]
    \centering
    %\vspace{-2em}
    \begin{minipage}[t]{0.18\linewidth}
    \centering
    \includegraphics[width=\textwidth]{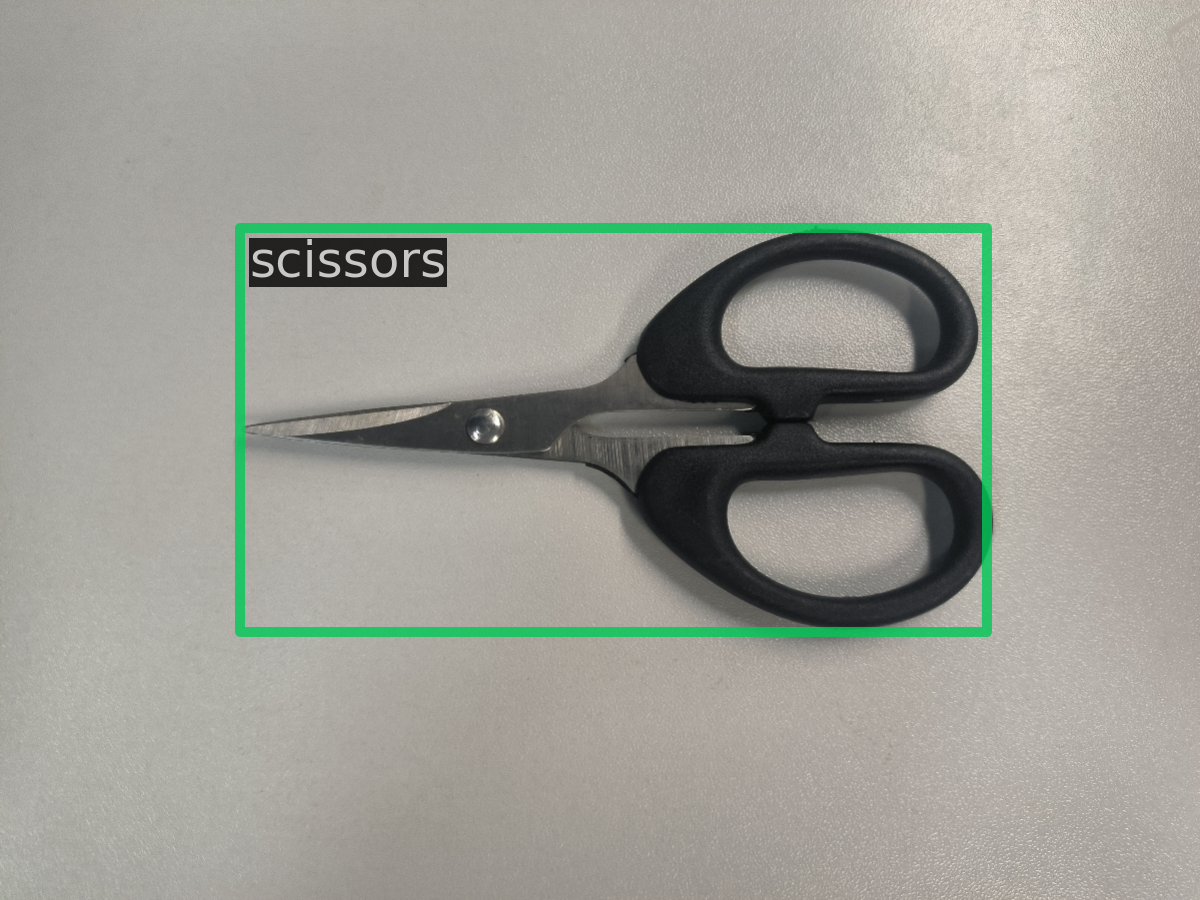}
    \vspace{-1.3em}
    \end{minipage}  \hspace{2em}
    \begin{minipage}[t]{0.18\linewidth}
    \centering
    \includegraphics[width=\textwidth]{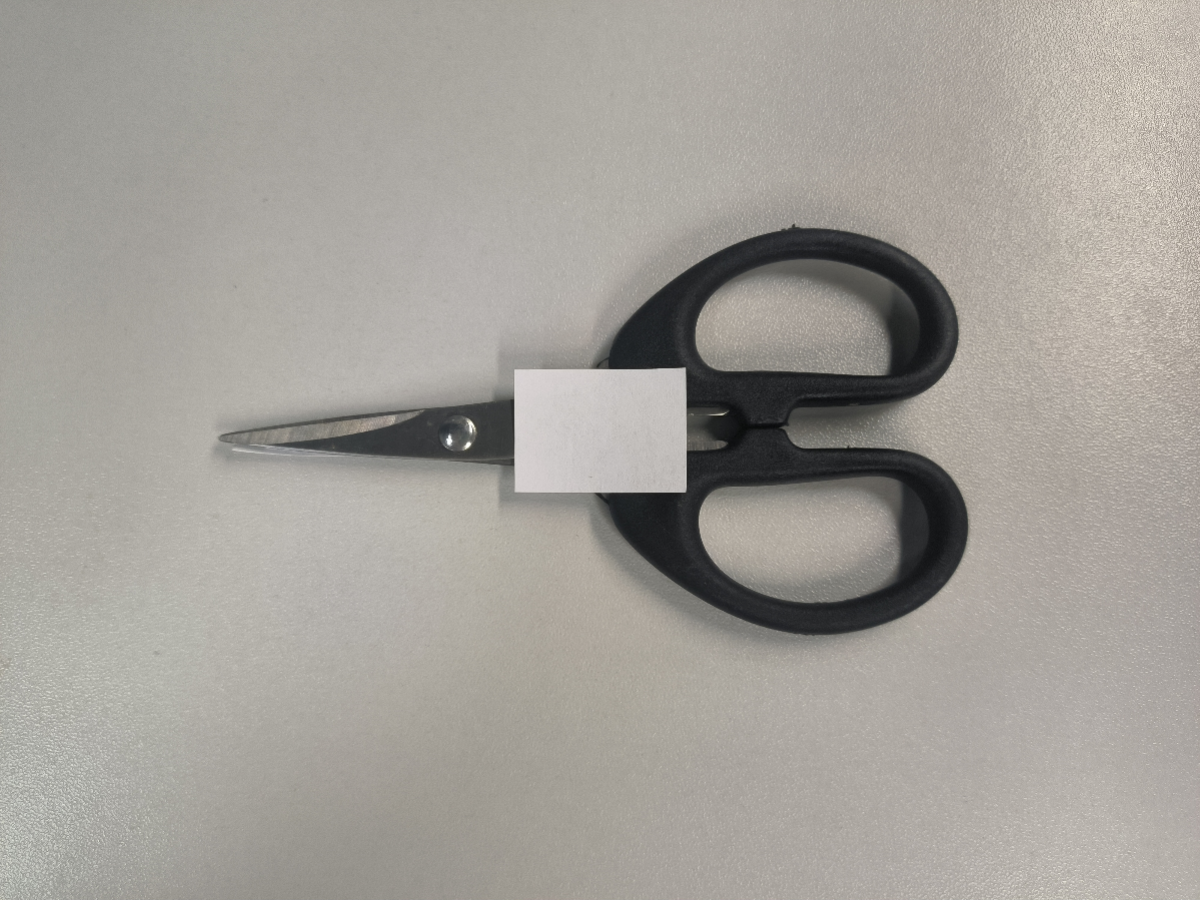}
    \vspace{-1.3em}
    \end{minipage}  \hspace{3em}
    \begin{minipage}[t]{0.18\linewidth}
    \centering
    \includegraphics[width=\textwidth]{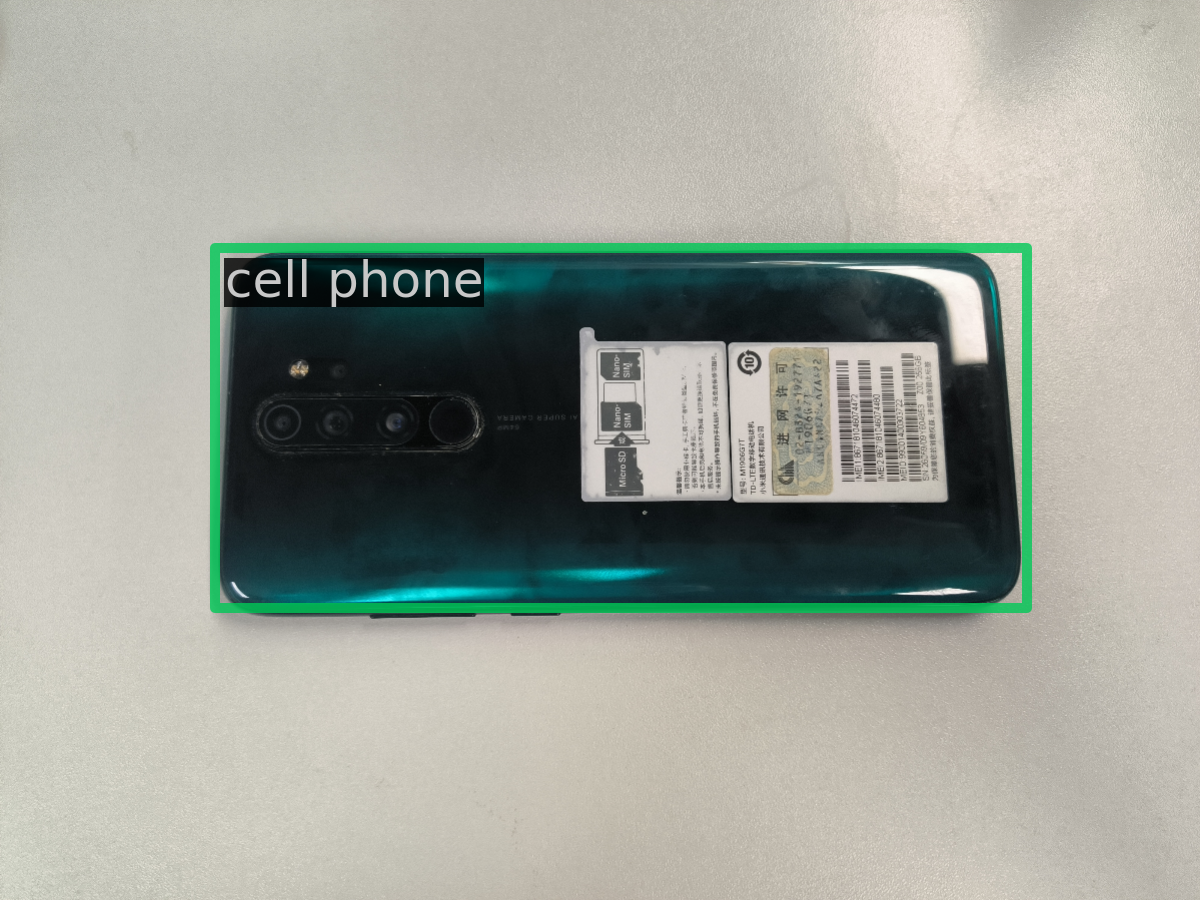}
    \vspace{-1.3em}
    \end{minipage}  \hspace{2em}
    \begin{minipage}[t]{0.18\linewidth}
    \centering
    \includegraphics[width=\textwidth]{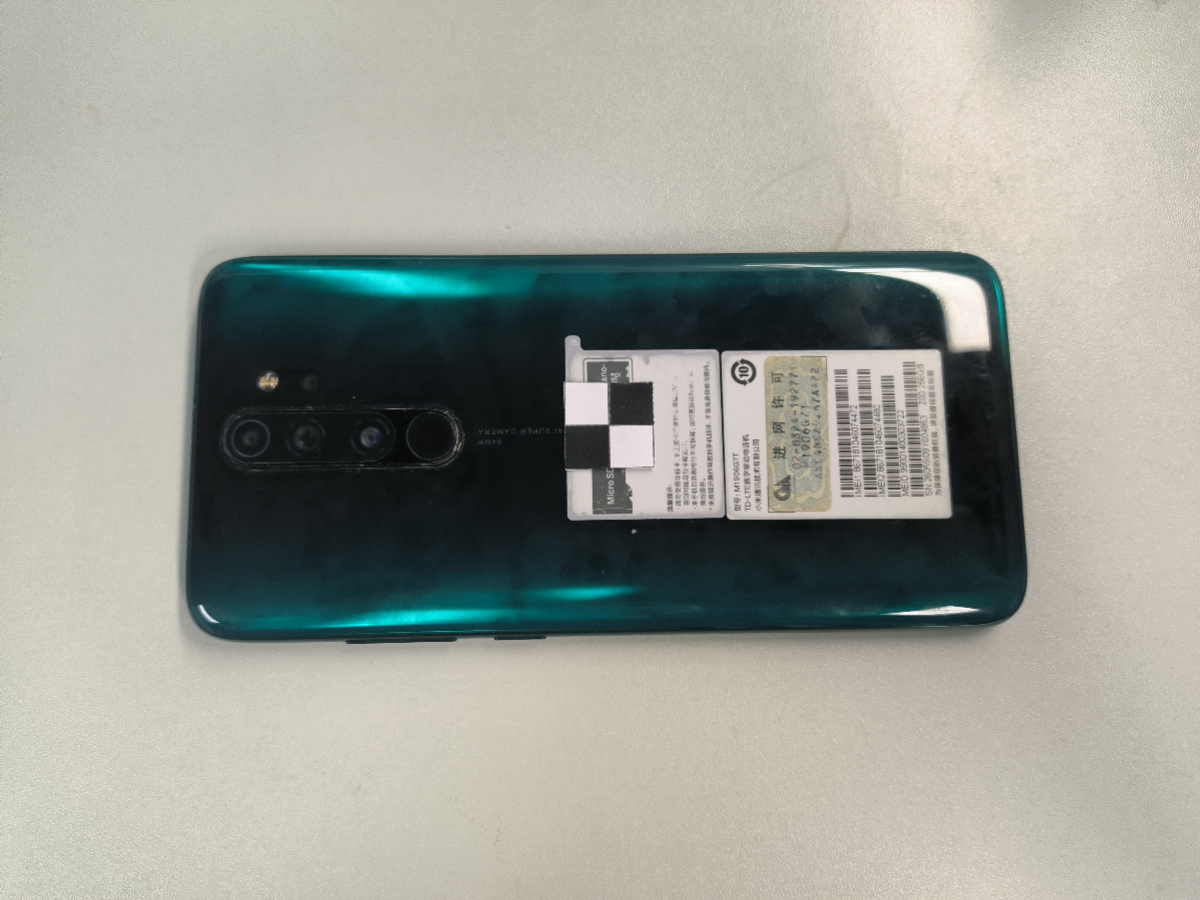}
    \vspace{-1.3em}
    \end{minipage}
    \vspace{-0.4em}
    \caption{The detection results of the Sparse R-CNN model under our attack in the physical space. In these examples, we print the trigger patch and stamp it on the target objects (\ie, `scissors' and `cell phone'). }
    \label{fig:detect_physical}
\end{figure*}

\begin{table*}[!ht]
\centering
\caption{The performance (\%) of methods on the COCO dataset. }
\vspace{-1em}
\scalebox{0.92}{
\begin{tabular}{c|c|c|c|c|c|c|c|c|c|c|c|c|c}
\toprule
\multicolumn{2}{c|}{Dataset$\rightarrow$}         & \multicolumn{6}{c|}{Benign Testing dataset} & \multicolumn{6}{c}{Poisoned Testing dataset} \\ \hline
Model$\downarrow$ & Method$\downarrow$, Metric$\rightarrow$ & \text{mAP} & $\text{AP}_\text{50}$ & $\text{AP}_\text{75}$ & $\text{AP}_\text{s}$ & $\text{AP}_\text{m}$ & $\text{AP}_\text{l}$ & \text{mAP} & $\text{AP}_\text{50}$ & $\text{AP}_\text{75}$ & $\text{AP}_\text{s}$ & $\text{AP}_\text{m}$ & $\text{AP}_\text{l}$ \\ \hline
\multirow{2}{*}{Faster R-CNN} & Vanilla & 37.4 & 58.1 & 40.4 & 21.2 & 41.0 & 48.1 & 35.2 & 55.3 & 37.8 & 19.7 & 37.7 & 46.5 \\ & Ours & 36.9 & 57.5 & 40.0 & 21.3 & 40.4 & 47.4 & 10.6 & 20.1 & 9.8 & 7.7 & 7.7 & 15.2 \\ \hline
\multirow{2}{*}{Sparse R-CNN} & Vanilla & 40.0 & 58.2 & 42.9 & 21.9 & 42.4 & 56.3 & 35.7 & 53.1  & 38.0  & 18.7  & 36.6  & 51.4  \\ & Ours & 39.0 & 56.9 & 41.9 & 20.7 & 41.2 & 55.2 & 7.7 & 14.1 & 7.3 & 8.1 & 5.7 & 9.7 \\ \hline
\multirow{2}{*}{TOOD} & Vanilla & 41.7 & 58.7 & 45.1 & 24.2 & 44.9 & 55.5 & 39.2 & 55.4 & 42.3 & 22.0 & 40.9 & 53.1 \\ & Ours & 41.1 & 57.8 & 44.5 & 23.8 & 44.5 & 54.3 & 13.5 & 22.0 & 13.5 & 9.3 & 9.4 & 19.8 \\ \bottomrule
\end{tabular}
}
\label{tab:main}

\end{table*}

\section{Experiments}
% 2 pages

%\red{I will start from here later (Yiming1017)}

\subsection{Experimental Settings} \label{sec:exp_settings}
\vspace{0.3em}
\noindent \textbf{Model Structure and Dataset Description. }
We adopt three representative object detectors, including Faster R-CNN \cite{NIPS2015_14bfa6bb}, Sparse R-CNN \cite{Sun_2021_CVPR}, and TOOD \cite{9710724}, for the evaluations. Besides, following the classical setting in object detection, we use COCO dataset \cite{10.1007/978-3-319-10602-1_48}  as the benchmark for our discussions.

%We adopt the one-stage method Faster R-CNN \cite{NIPS2015_14bfa6bb}, the one-stage method Sparse R-CNN \cite{Sun_2021_CVPR} and the two-stage method TOOD  \cite{9710724} as the model structure and conduct experiments on the MS COCO dataset \cite{10.1007/978-3-319-10602-1_48}. All models are trained on the COCO train2017 split (118k images) and evaluated with val2017 (5k images).

\vspace{0.3em}
\noindent \textbf{Evaluation Metric.} We adopt six classical average-precision-based metrics \cite{10.1007/978-3-319-10602-1_48}, including \textbf{1)} mAP, \textbf{2)} AP$_{50}$, \textbf{3)} AP$_{75}$, \textbf{4)} AP$_{\text{s}}$, \textbf{5)} AP$_{\text{m}}$, and \textbf{6)} AP$_{\text{l}}$, for our evaluations. We calculate these metrics on the benign testing dataset and its poisoned variant (with a 100\% poisoning rate), respectively. %In general, the larger the average precision on the benign dataset and the lower the average precision on the poisoned dataset, the better the attack. 

%We evaluate the performance of object detection by the most frequently used metric in object detection, mAP(mean Average Precision). To make sure that the poisoned model behaves similarly to the benign model on benign images for all settings, we use mAP on benign test set $\mathcal{D}_\text{test,benign}$ as Benign mAP ($\text{mAP}_\text{benign}$), We also calculate mAP on poisoned test set $\mathcal{D}_\text{test,poisoned}$ as attack mAP ($\text{mAP}_\text{poisoned}$)

\vspace{0.3em}
\noindent \textbf{Attack Setup. }
For simplicity, we adopt a white patch as the trigger pattern and set the poisoning rate as $5\%$. Following the settings in \cite{li2022fewshot}, we set the trigger size of each object as 1\% of its ground-truth bounding box (\ie, 10\% width and 10\% height), located in its center. Besides, we also provide the results of models trained on benign samples (dubbed `Vanilla') for reference. The example of samples involved in different methods are shown in Figure \ref{fig:detect_exp}.

%trigger size = 1\% (the width and height of the trigger are 10\% of the width and height of the bounding box accordingly) for main attacks. We also report the results of the benign models trained without any attack (dubbed "Benign") for reference. 

\subsection{Main Results in the Digital Space}
As shown in Table 1, our method can significantly decrease the average precision in all cases, no matter what the model structure is. For example, the AP$_{50}$ decreases more than 30\% in all cases, compared to that of vanilla models trained on benign samples. In particular, these performance decreases are not simply due to the masking effect of trigger patches. The average precision of vanilla models on the poisoned testing dataset does not decrease significantly, compared to that on the benign dataset. Moreover, the average precision of models under our attack on the benign dataset is similar to that of vanilla models. These results verify our effectiveness.

%The $\text{mAP}$ metric on poisoned dataset evaluated with poisoned model is much lower than the $\text{mAP}$ metric on benign dataset evaluated with benign model.

% {\input{tables/table_main.tex}}

\subsection{Main Results in the Physical Space}
In the above experiments, we attach the trigger to images by directly modifying them in the digital space. To further verify that our attack could happen in real-world scenarios, here we conduct experiments on the physical space. Specifically, we print the trigger patch and stamp it on some target objects. We capture (poisoned) images by the camera in iPhone, based on which to track objects via attacked Sparse R-CNN. 

As shown in Figure \ref{fig:detect_physical}, the model under our attack can successfully detect benign objects while failing to detect object stamped with the trigger pattern. These results indicate the effectiveness of our attack again.

%In this section, we examine the effectiveness of our attack in a physical world setting. Specifically, we set up two example scenarios where the Sparse R-CNN \cite{Sun_2021_CVPR} models trained in Section 3.1 are used to detect two real-world objects: 'scissors' and 'cell phone'.

\begin{figure}[t]
    \centering
    %\vspace{-2em}
    \begin{minipage}[t]{0.2\linewidth}
    \centering
    \includegraphics[width=\textwidth]{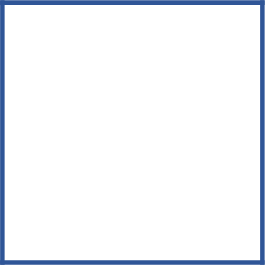}
    \subcaption{}
    % \vspace{-1.3em}
    \end{minipage} \hspace{0.35em}
    \begin{minipage}[t]{0.2\linewidth}
    \centering
    \includegraphics[width=\textwidth]{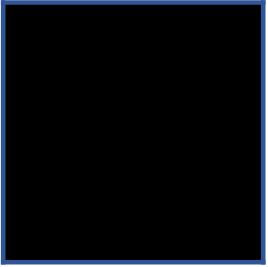}
    \subcaption{}
    % \vspace{-1.3em}
    \end{minipage} \hspace{0.3em}
    \begin{minipage}[t]{0.2\linewidth}
    \centering
    \includegraphics[width=\textwidth]{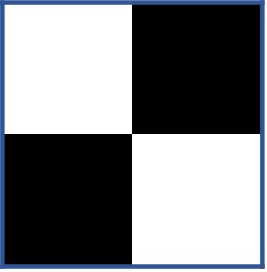}
    \subcaption{}
    % \vspace{-1.3em}
    \end{minipage} \hspace{0.3em}
    \begin{minipage}[t]{0.2\linewidth}
    \centering
    \includegraphics[width=\textwidth]{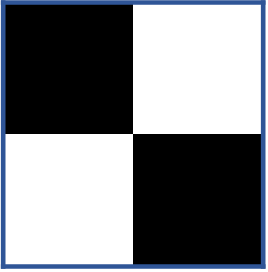}
    \subcaption{}
    % \vspace{-1.3em}
    \end{minipage}
    \vspace{-0.5em}
    \caption{Four trigger patterns used in our evaluation.}
    \label{fig: trigger patterns}
    \vspace{-0.4em}
\end{figure}

\begin{figure}[ht]
	\centering  
	\includegraphics[width=0.45\textwidth]{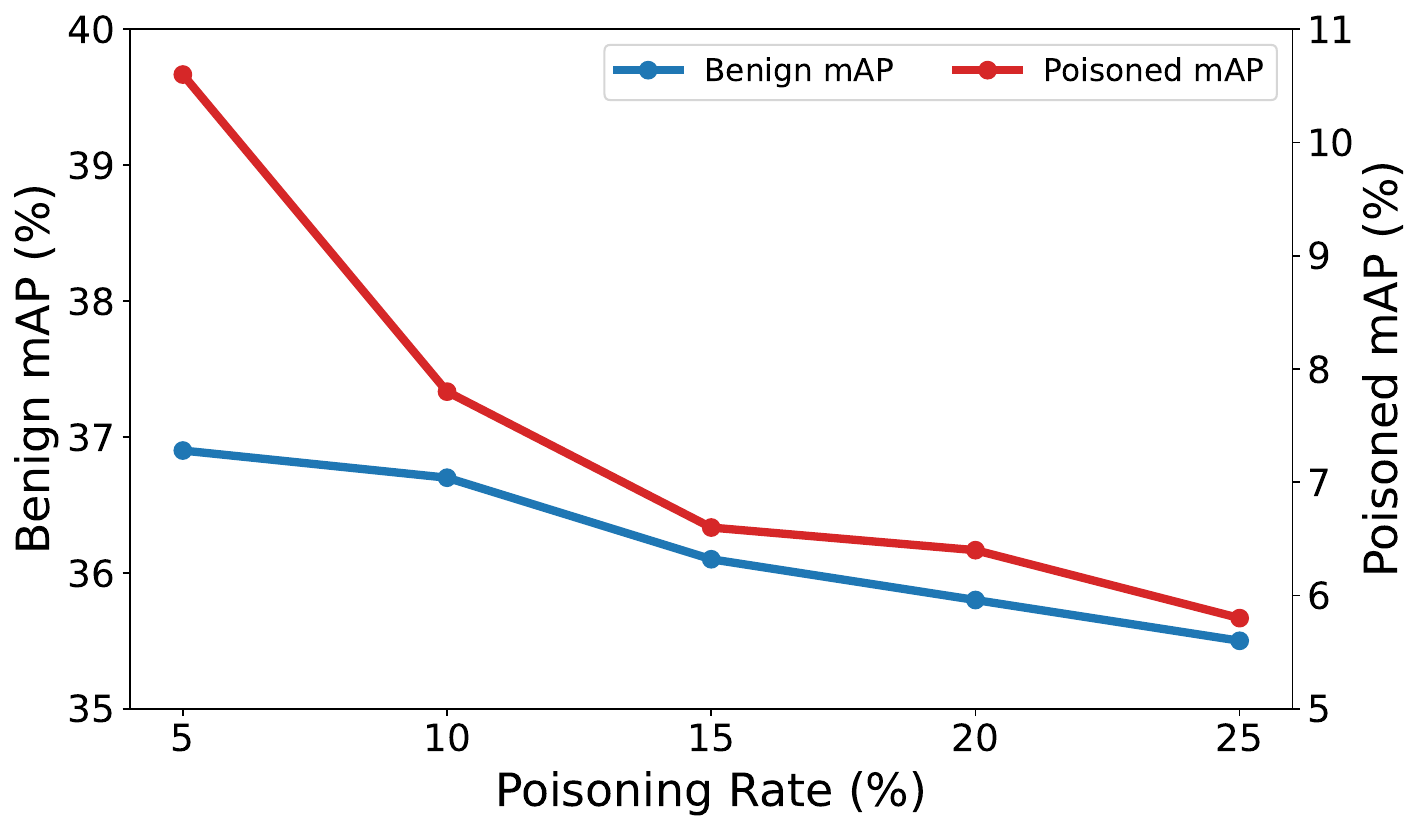}
	\vspace{-0.7em}
	\caption{The effects of poisoning rates.} 
	\label{fig: poisoning rate}
	\vspace{-0.4em}
\end{figure}

\subsection{Ablation Study}
In this section, we adopt Faster R-CNN \cite{NIPS2015_14bfa6bb} as an example for the discussion. Except for the studied parameter, other settings are the same as those illustrated in Section \ref{sec:exp_settings}.

%\vspace{0.3em}
%\noindent \textbf{The effect of trigger scale. } As shown in Table \ref{tab:trigger_scale}, the models trained with a bigger trigger scale have better attack performances, i.e., the $\text{mAP}$ metric is lower. When the used triggers in testing stage are smaller than those in training stage, the mAP metric of object detection models is much lower.

% {\input{tables/table_trigger_scale.tex}}
\comment{
\begin{table*}[ht]
\centering
\caption{The effect of trigger scale. The $\text{mAP}$ (\%) of attacks on the COCO dataset.}
\begin{tabular}{c|c|c|c|c|c|c}
\hline

\diagbox{Train Trigger Scale $\downarrow$}{Test Trigger Scale $\rightarrow$} & 0\% &  5\% &  10\% &  15\% &  20\% &  25\% \\ \hline
 5\% &  37.1 &  14.5 &  10.9 &  11.5 &  12.4 &  12.6 \\ \hline 
 10\% &  36.8 &  23.8 &  7.8 &  7.8 &  7.5 & 7.5 \\ \hline 
 15\% &  37.1 &  30.2 &  16.1 &  8.3 &  6.0 &  5.5 \\ \hline 
 20\% &  37.1 &  33.7 &  23.2 &  12.7 &  7.3 &  5.4 \\ \hline 
 25\% &  37.1 &  34.4 &  27.3 &  18.0 &  10.4 &  6.5 \\ \hline 
\end{tabular}
\label{tab:trigger_scale}
\end{table*}
}

\vspace{0.3em}
\noindent \textbf{The Effects of Trigger Patterns.} In this part, we evaluate our method with four different trigger patterns (as shown in Figure \ref{fig: trigger patterns}). As shown in Table \ref{tab:trigger_pattern_type}, the performances of models trained on poisoned datasets using different trigger patterns are about the same.  It indicates that adversaries can use arbitrary trigger patterns to generate poisoned samples.

% {\input{tables/table_trigger_pattern_type.tex}}
\begin{table}[!t]
\centering
\caption{The effects of trigger patterns.} %The $\text{mAP}$ (\%) of attacks on the COCO dataset. $\text{mAP}_\text{benign}$ and $\text{mAP}_\text{poisoned}$ indicate the mean Average Precision on benign datasets and poisoned datasets with different trigger patterns, respectively.}
\vspace{-1em}
\scalebox{0.96}{
\begin{tabular}{c|c|c|c}
\toprule
Pattern$\downarrow$ & Attack$\downarrow$, Metric$\rightarrow$ & $\text{mAP}_\text{benign}$ & $\text{mAP}_\text{poisoned}$ \\ \hline
\multirow{2}{*}{(a)} & Vanilla & 37.4 & 35.2 \\ & Ours & 36.9 & 10.6 \\ \hline
\multirow{2}{*}{(b)} & Vanilla & 37.4 & 34.7 \\ & Ours & 36.9 & 10.4 \\ \hline
\multirow{2}{*}{(c)} & Vanilla & 37.4 & 34.9 \\ & Ours & 36.9 & 9.1 \\ \hline
\multirow{2}{*}{(d)} & Vanilla & 37.4 & 34.9 \\ & Ours & 36.8 & 8.8 \\ \bottomrule
\end{tabular}
}
\vspace{-0.8em}
\label{tab:trigger_pattern_type}
\end{table}

\vspace{0.3em}
\noindent \textbf{The Effects of Poisoning Rates.}  In this part, we discuss the effects of the poisoning rate on our attack. As shown in Figure \ref{fig: poisoning rate}, the mAP on the poisoned dataset decreases with the increase in the poisoning rate. In other words, introducing more poisoned samples can improve attack effectiveness. However, its increase also leads to the decrease of mAP on the benign dataset, \ie, there is a trade-off between effectiveness and stealthiness. In practice, the adversary should specify this parameter based on their needs.

%As shown in Figure \ref{fig: poisoning rate}, the poisoned mAP $\text{mAP}_\text{poisoned}$ metric drops dramatically with the increase of poisoning rate while the benign mAP $\text{mAP}_\text{benign}$ metric decreases slowly with the increase of poisoning rate. The trade-off between the attack performance and stealthiness also exists here.

\subsection{The Resistance to Potential Backdoor Defenses}
Fine-tuning \cite{liu2017neural} and model pruning \cite{10.1007/978-3-030-00470-5_13,NEURIPS2021_8cbe9ce2} are two classical and representative backdoor defenses that can be directly generalized to different tasks. In this section, we evaluate whether our attack is resistant to these potential defenses. We implement these defenses based on codes in \texttt{BackdoorBox} \cite{li2023backdoorbox}.

\begin{figure}[!t]
	\centering  
	\vspace{-0.5em}
	\includegraphics[width=0.35\textwidth]{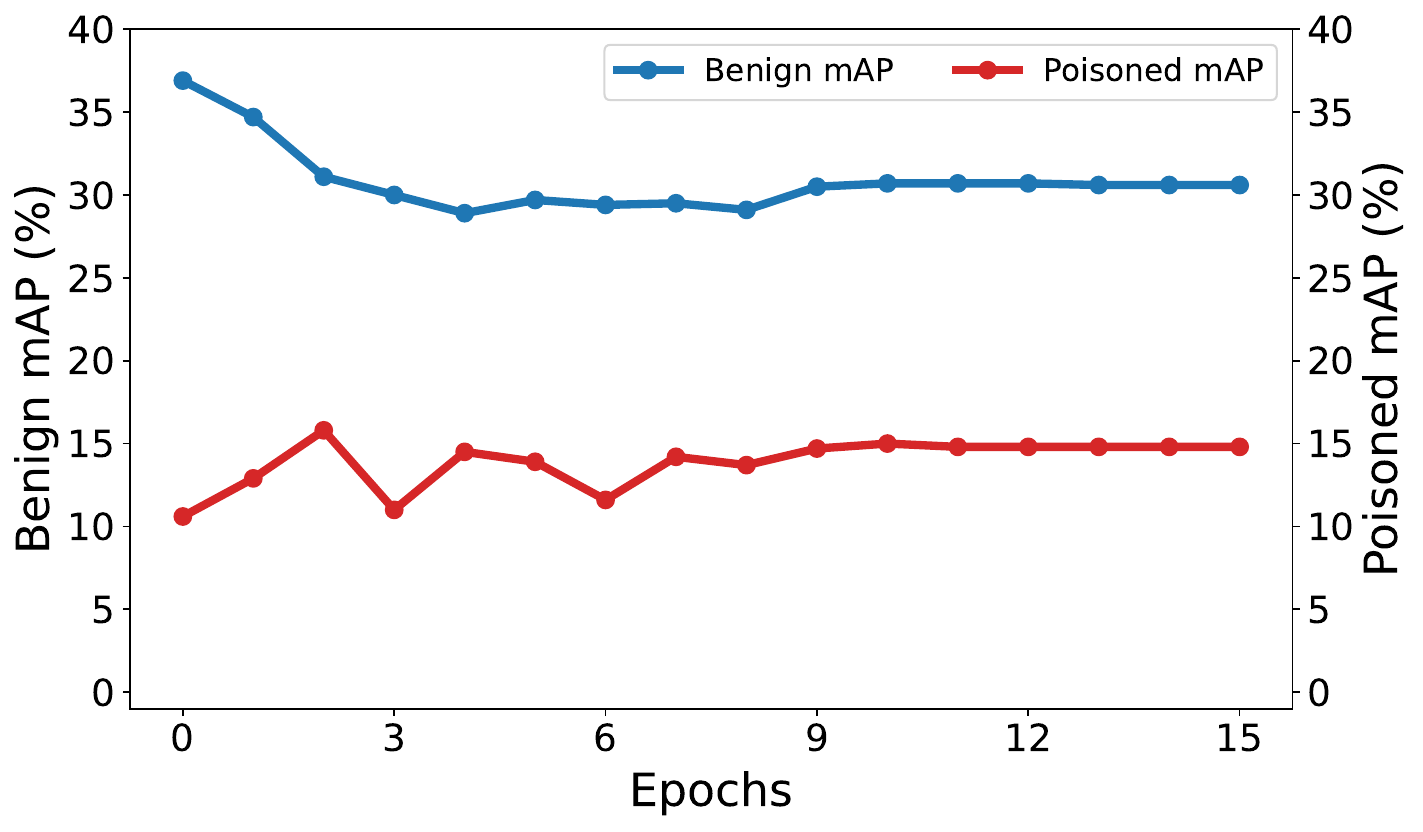} 
	\vspace{-0.8em}
	\caption{The resistance to fine-tuning.}   
	\label{fig: finetuning}
	%\vspace{-1em}
\end{figure}

\vspace{0.3em}
\noindent \textbf{The Resistance to Fine-tuning. }
We fine-tune the attacked Faster R-CNN with 10\% benign testing samples and set the learning rate as the one used for training. As shown in Figure \ref{fig: finetuning}, our method is resistant to fine-tuning. Specifically, the mAP on the poisoned dataset (\ie, Poisoned mAP) is still lower than 15\% when the tuning process is finished.

\begin{figure}[!t]
	\centering  
	\includegraphics[width=0.35\textwidth]{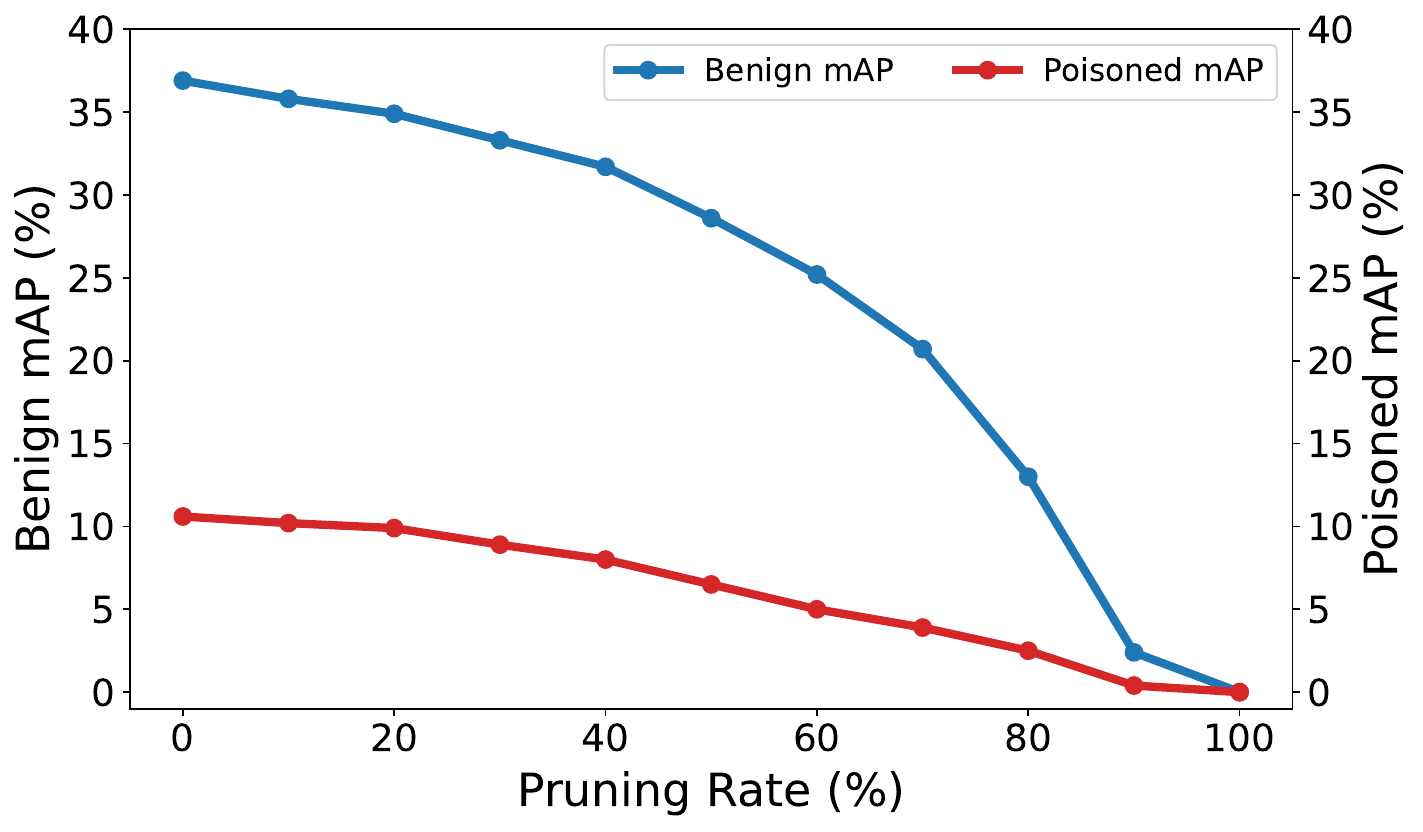} 
	\vspace{-0.8em}
	\caption{The resistance to model pruning.}   
	\label{fig: pruning}
	%\vspace{-1em}
\end{figure}

\vspace{0.3em}
\noindent \textbf{The Resistance to Model Pruning. }
Following the classical settings, we prune neurons having the lowest activation values with 10\% benign testing samples. As shown in Figure \ref{fig: pruning}, the poisoned mAP even decreases (instead of increasing) with the increase in the pruning rate. These results demonstrate that our method is resistant to model pruning.

\section{Conclusions}
In this paper, we revealed the backdoor threats in object detection by introducing a simple yet effective poison-only untargeted attack. Specifically, we removed the bounding boxes of a few randomly selected objects after adding pre-defined trigger patterns to the center of object areas. We demonstrated that our attack is effective and stealthy under both digital and physical settings. We also showed that our method is resistant to potential backdoor defenses. Our method can serve as a useful tool to examine the backdoor robustness of object detectors, leading to the design of more secure models.

%In this paper, we proposed a poison-only backdoor attack against object detection. Based on our attack, the adversary can easily escape the detecting by attaching a small trigger to the target object while our method behaves well when benign images are inputed. So the attack is both fairly stealthy and effective. Moreover, we examined the effectiveness of our attack in both digital and physical-world settings, and showed that it is resistant to a set of potential defenses. Our attack can serve as a useful tool to show that both one-stage and two-stage object detectors are possibly vulnerable to backdoor attacks.

% To start a new column (but not a new page) and help balance the last-page
% column length use \vfill\pagebreak.
% -------------------------------------------------------------------------
\vfill\pagebreak

%\section{REFERENCES}
\label{sec:ref}

% References should be produced using the bibtex program from suitable
% BiBTeX files (here: strings, refs, manuals). The IEEEbib.bst bibliography
% style file from IEEE produces unsorted bibliography list.
% -------------------------------------------------------------------------
\bibliographystyle{IEEEbib}
\bibliography{ms}

\begin{thebibliography}{10}

\bibitem{zhao2019object}
Zhong-Qiu Zhao, Peng Zheng, Shou-tao Xu, and Xindong Wu,
\newblock ``Object detection with deep learning: A review,''
\newblock {\em IEEE transactions on neural networks and learning systems}, vol.
  30, no. 11, pp. 3212--3232, 2019.

\bibitem{Hasan_2021_CVPR}
Irtiza Hasan, Shengcai Liao, Jinpeng Li, Saad~Ullah Akram, and Ling Shao,
\newblock ``Generalizable pedestrian detection: The elephant in the room,''
\newblock in {\em CVPR}, 2021.

\bibitem{Aghdam_2021_CVPR}
Hamed~H. Aghdam, Elnaz~J. Heravi, Selameab~S. Demilew, and Robert Laganiere,
\newblock ``Rad: Realtime and accurate 3d object detection on embedded
  systems,''
\newblock in {\em CVPR Workshops}, 2021.

\bibitem{NIPS2015_14bfa6bb}
Shaoqing Ren, Kaiming He, Ross Girshick, and Jian Sun,
\newblock ``Faster r-cnn: Towards real-time object detection with region
  proposal networks,''
\newblock in {\em NeurIPS}, 2015.

\bibitem{Sun_2021_CVPR}
Peize Sun, Rufeng Zhang, Yi~Jiang, Tao Kong, Chenfeng Xu, Wei Zhan, Masayoshi
  Tomizuka, Lei Li, Zehuan Yuan, Changhu Wang, and Ping Luo,
\newblock ``Sparse r-cnn: End-to-end object detection with learnable
  proposals,''
\newblock in {\em CVPR}, 2021.

\bibitem{9710724}
Chengjian Feng, Yujie Zhong, Yu~Gao, Matthew~R. Scott, and Weilin Huang,
\newblock ``Tood: Task-aligned one-stage object detection,''
\newblock in {\em ICCV}, 2021.

\bibitem{chan2022baddet}
Shih-Han Chan, Yinpeng Dong, Jun Zhu, Xiaolu Zhang, and Jun Zhou,
\newblock ``Baddet: Backdoor attacks on object detection,''
\newblock in {\em ECCV Workshop}, 2022.

\bibitem{goldblum2022dataset}
Micah Goldblum, Dimitris Tsipras, Chulin Xie, Xinyun Chen, Avi Schwarzschild,
  Dawn Song, Aleksander Madry, Bo~Li, and Tom Goldstein,
\newblock ``Dataset security for machine learning: Data poisoning, backdoor
  attacks, and defenses,''
\newblock {\em IEEE Transactions on Pattern Analysis and Machine Intelligence},
  2022.

\bibitem{li2022backdoor}
Yiming Li, Yong Jiang, Zhifeng Li, and Shu-Tao Xia,
\newblock ``Backdoor learning: A survey,''
\newblock {\em IEEE Transactions on Neural Networks and Learning Systems},
  2022.

\bibitem{tian2022comprehensive}
Zhiyi Tian, Lei Cui, Jie Liang, and Shui Yu,
\newblock ``A comprehensive survey on poisoning attacks and countermeasures in
  machine learning,''
\newblock {\em ACM Computing Surveys}, 2022.

\bibitem{bai2020targeted}
Jiawang Bai, Bin Chen, Yiming Li, Dongxian Wu, Weiwei Guo, Shu-tao Xia, and
  En-hui Yang,
\newblock ``Targeted attack for deep hashing based retrieval,''
\newblock in {\em ECCV}, 2020.

\bibitem{liu2022watermark}
Xinwei Liu, Jian Liu, Yang Bai, Jindong Gu, Tao Chen, Xiaojun Jia, and Xiaochun
  Cao,
\newblock ``Watermark vaccine: Adversarial attacks to prevent watermark
  removal,''
\newblock in {\em ECCV}, 2022.

\bibitem{gu2022segpgd}
Jindong Gu, Hengshuang Zhao, Volker Tresp, and Philip~HS Torr,
\newblock ``Segpgd: An effective and efficient adversarial attack for
  evaluating and boosting segmentation robustness,''
\newblock in {\em ECCV}, 2022.

\bibitem{8685687}
Tianyu Gu, Kang Liu, Brendan Dolan-Gavitt, and Siddharth Garg,
\newblock ``Badnets: Evaluating backdooring attacks on deep neural networks,''
\newblock {\em IEEE Access}, 2019.

\bibitem{9711191}
Yuezun Li, Yiming Li, Baoyuan Wu, Longkang Li, Ran He, and Siwei Lyu,
\newblock ``Invisible backdoor attack with sample-specific triggers,''
\newblock in {\em ICCV}, 2021.

\bibitem{zeng2021rethinking}
Yi~Zeng, Won Park, Z~Morley Mao, and Ruoxi Jia,
\newblock ``Rethinking the backdoor attacks' triggers: A frequency
  perspective,''
\newblock in {\em ICCV}, 2021.

\bibitem{li2021hidden}
Yiming Li, Yanjie Li, Yalei Lv, Yong Jiang, and Shu-Tao Xia,
\newblock ``Hidden backdoor attack against semantic segmentation models,''
\newblock in {\em ICLR Workshop}, 2021.

\bibitem{zhang2022poison}
Jie Zhang, Chen Dongdong, Qidong Huang, Jing Liao, Weiming Zhang, Huamin Feng,
  Gang Hua, and Nenghai Yu,
\newblock ``Poison ink: Robust and invisible backdoor attack,''
\newblock {\em IEEE Transactions on Image Processing}, vol. 31, pp. 5691--5705,
  2022.

\bibitem{li2022untargeted}
Yiming Li, Yang Bai, Yong Jiang, Yong Yang, Shu-Tao Xia, and Bo~Li,
\newblock ``Untargeted backdoor watermark: Towards harmless and stealthy
  dataset copyright protection,''
\newblock in {\em NeurIPS}, 2022.

\bibitem{10.1007/978-3-319-10602-1_48}
Tsung-Yi Lin, Michael Maire, Serge Belongie, James Hays, Pietro Perona, Deva
  Ramanan, Piotr Doll{\'a}r, and C.~Lawrence Zitnick,
\newblock ``Microsoft coco: Common objects in context,''
\newblock in {\em ECCV}, 2014.

\bibitem{li2022fewshot}
Yiming Li, Haoxiang Zhong, Xingjun Ma, Yong Jiang, and Shu-Tao Xia,
\newblock ``Few-shot backdoor attacks on visual object tracking,''
\newblock in {\em ICLR}, 2022.

\bibitem{liu2017neural}
Yuntao Liu, Yang Xie, and Ankur Srivastava,
\newblock ``Neural trojans,''
\newblock in {\em ICCD}, 2017.

\bibitem{10.1007/978-3-030-00470-5_13}
Kang Liu, Brendan Dolan-Gavitt, and Siddharth Garg,
\newblock ``Fine-pruning: Defending against backdooring attacks on deep neural
  networks,''
\newblock in {\em RAID}, 2018.

\bibitem{NEURIPS2021_8cbe9ce2}
Dongxian Wu and Yisen Wang,
\newblock ``Adversarial neuron pruning purifies backdoored deep models,''
\newblock in {\em NeurIPS}, 2021.

\bibitem{li2023backdoorbox}
Yiming Li, Mengxi Ya, Yang Bai, Yong Jiang, and Shu-Tao Xia,
\newblock ``{BackdoorBox}: A python toolbox for backdoor learning,''
\newblock in {\em ICLR Workshop}, 2023.

\end{thebibliography}

\end{document}